\documentclass{article}
\usepackage{spconf,amsmath,graphicx}
\usepackage[noend,linesnumbered,ruled,vlined]{algorithm2e}
\usepackage{multirow}

\title{DFMU: Data-Frugal Machine Unlearning}
\name{Sajith U, Prateek Keserwani}
\address{Samsung R\&D Institute India-Bangalore, India\\ \{sajith.u, k.prateek\}@samsung.com}

\begin{document}
	%
	\maketitle
	\begin{abstract}
		Machine unlearning is an emerging domain that ensures the safe removal of elements (includes concepts, attributes, entity and class) from the trained model along with least drop in model performance. The domain of machine unlearning brings its own indigenous challenges since the removal of pre-trained elements from model will always degrade the model performance on remaining elements. The existing methods basically rely on retraining for removal of elements
		from the pre-trained model, which is compute extensive. In this work, we propose a machine unlearning method which helps to reduce the computational requirement for faster retain-dataset accuracy convergence which also does not require extensive retraining of the pre-trained model. The proposed method, Data-Frugal Machine Unlearning (DFMU) requires only a single forward and backward pass for computing the importance score of various computational blocks of a model. The importance score computation
		is based on knowledge preserving pruning which helps to converge faster and requires far less data as compared to the existing methods. Experimentally, it achieves 40\% more retain-accuracy with just 13\% of data samples in comparison with SOTA method on various public datasets and also averages 88\% faster processing time for forgetting a given class.
	\end{abstract}
	\section{Introduction}
	\label{sec:intro}
	
	In recent years, machine learning has experienced a surge in popularity, driven by technological advancements, the increasing availability of data, and its transformative influence across numerous industries. As its applications continue to expand into a wide array of areas in everyday life, the regulatory requirements associated with it are also growing. These regulations often mandate the removal of specific types of data, tasks, categories, or concepts (e.g., private, sensitive, or harmful information) from trained models to ensure compliance. In the light of conforming to compliance norms, forgetting or removal of the information contained in a machine learning model becomes an important topic of discussion. The ability of a model to remove certain identified data points (forget set) and its influences from a trained model along with retain rest data points intact (retain set) is termed as Machine Unlearning \cite{li2025machine}
	
	The existing training-based unlearning solutions are based on first-order gradients to identify the potential neurons that can be targeted to utilize for removal of concepts, tasks, or categories. These methods needs the entire dataset to be preserved which may not be practical every time. To counter this challenge, retraining-free methods has introduced which depends on finding the Fisher Information Matrix \cite{guo2019certified}. It makes algorithm of machine unlearning computationally expensive. 
	
	\begin{figure}[!t]
		\centering
		\includegraphics[width=0.5\textwidth]{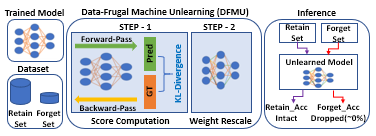}
		\caption{Illustration of Data-Frugal Machine Unlearning}
		\label{fig::overview}
	\end{figure}	
	
	In order to eliminate the drawback of above mentioned priors, we have taken inspiration from the zero-cost proxies of neural architecture search \cite{abdelfattah2021zero,keserwani2023receptive}. In this work, we propose a data-frugal machine unlearning algorithm (DFMU) designed to preserve the knowledge of the retain set which at the same time remove elements from forget set efficiently. To achieve this, we have utilized the knowledge preserving pruning algorithm \cite{park2023accurate}. It measures the knowledge present in each layer of a trained model prior to computing the importance scores based on the forget and complete dataset. Further, using the importance scores as a guide, the model weights are rescaled such that the forget information is removed and the remaining data information is kept intact in the model which don’t requires retraining. The proposed idea is depicted in Fig. \ref{fig::overview}. 
	
	\section{Related Work}
	\label{sec:literature}
	Machine unlearning methods are composed of identification of the target parameters and followed by  modifying/nullifying the target parameters. These two tasks work together to generate a model which has unlearned the targeted information. Machine unlearning works are broadly categorized as retraining-based \cite{chundawat2023can,tarun2023deep} and retraining-free machine unlearning \cite{mehta2022deep, foster2024fast,foster2024loss, jia2023model}. In \cite{chundawat2023can}, a teacher-student framework is used with competent and incompetent teachers and the student unlearned model is obtained via selective knowledge transfer. In \cite{tarun2023deep} two deep regression unlearning methods are introduced namely Blindspot learning and Gaussian Amnesiac learning for regression tasks. In comparison, the proposed work is restricted to retraining-free machine unlearning.   
	
	Retraining-free methods attempts to achieve machine unlearning without having to retrain or with less training on retained data.  In \cite{mehta2022deep}, a subset  of parameters is picked based on conditional independence coefficient to unlearn and removes the need to modify the whole parameters of model. It helps to reduce the training cost. In \cite{foster2024fast} a post-hoc retrain-free method is introduced. It uses fisher information matrix of training and forget set to identify parameters that are more important for forget set. These parameters are proportionally dampened so that the model's dependency on forget data is reduced. This work is extended by \cite{foster2024loss} by introducing a method that is retraining-free as well as label-free. It uses approximation based on model output sensitivity rather than Fisher information.  In \cite{jia2023model} the concept of pruning is used to approximate machine unlearning. It first sparsify the model via pruning so that the update to remove influence are cheaper and easier. While pruning and unlearnig requires some fine-tuning but reduces the number of parameters. Compared with the above mentioned retraining-free methods, we have adopted the zero-cost proxies~\cite{abdelfattah2021zero} approach used in Neural Architecture Search to further reduce the computational cost. It helps to converge fast. 
	
	\section{Proposed Method}
	\label{sec:format}
	
	The proposed approach calculates the importance score of computing blocks of a model with respect to the forget-dataset ($D_{f}$) and the complete-dataset ($D_{o}= D_{f}+D_{r}$). These importance scores are utilized to rescale the weights of selected layers, by which the model’s knowledge about the forget-dataset is removed and the remaining knowledge remains intact for the retain set ($D_{r}$), where  ($D_{o}= D_{f}+D_{r}$).

	\begin{algorithm}[!t]
		\SetAlgoLined
		\SetKwInOut{Input}{Input}
		\SetKwInOut{Output}{Output}
		\Input{Forget Dataset ($D_f$), pretrained model ($M$)}
		\Output{Importance Score ($s$) for $D_f$}
		$\hat{Y} \gets M(D_f)$ \\
		$L$ $\gets$ $D_{f_{KL}}(Y \parallel \hat{Y})$ \\
		Backpropagation wrt $L$
		
		\ForEach{layer $l$ in $N$}{
			$f$ $\gets$ get features(l) \\
			$w$ $\gets$ get weight(l) \\
			$s$ $\gets$ $\sum_{}^{}(f^2)\times \frac{\sum_{i=1}{n}w^2}{n}$	
		}
		\KwRet{$s$}
		\caption{Importance Scores Computation}
	\end{algorithm}	
	
	\begin{algorithm}[!t]
				\SetAlgoLined
		\SetKwInOut{Input}{Input}
		\SetKwInOut{Output}{Output}
		\Input{Importance scores ($S_{f}$) for $D_f$, pretrained model $M$, scaling const $\alpha$, perturb const $\lambda$}
		\Output{Model with unlearned weights ($M_{UL}$)}
		
		\ForEach{layer $l$ in $M$}{
			$\omega$ $\gets S_{0} \alpha / S_{f}$ \\
			
			$m \gets Zeros(size(\omega))$ \tcp{mask init} 
			
			$m \gets \begin{cases} 
				1 & \text{if } \omega < 1 \\
				0 & \text{otherwise} 
			\end{cases}$
			
			$W_{UL}[l] \gets W_{PT}[l] \times (m\omega \lambda)/\alpha$
		}
		\KwRet{$M_{UL}$}
		\caption{Weight Rescaling}
	\end{algorithm}

	\subsection{Importance Score Calculation}
	\label{ssec:subsubhead}
	
	The primary step for forgetting is the identification of salient computing blocks of the model, i.e. to identify those parameters which contribute to the learning of the forget-dataset and the complete-dataset. The importance score computation is inspired from the zero-cost proxies in neural architecture search~\cite{abdelfattah2021zero}. The single forward pass and backward pass for each data point with respect to the KL Divergence of predicted and ground truth is used to compute the importance score of computing block. This can be done for both forget set $D_f$ and complete dataset $D_o$. The importance scores $S_f$ and $S_o$ are computed by following the steps of Algorithm 1. The importance score is computed by using the knowledge preserving pruning method \cite{park2023accurate}. These knowledge preserving pruning score is utilized to setup a ground to identify the computing block sensitivity for retain and forget set. 
	
	\subsection{Weight Rescaling}
	\label{ssec:subsubhead}
	Once the importance scores are calculated, the next step is to apply weight scaling to the pre-trained model $M$, so that the intended class information will be removed, and the remaining knowledge of the model is retained. The importance score for forget set $S_{f}$ and complete set $S_{o}$ is used to rescale the initial model weights $W_{PT}$ and produce the unlearned weights $W_{UL}$. The detailed steps are described in Algorithm 2.
	
	\section{Experiments}
	\label{sec:majhead}
	
	\subsection{Experimental Setup}
	The proposed method is implemented on top of PyTorch~\cite{paszke2019pytorch}. Effectiveness of this approach is evaluated using Vision Transformer (ViT)~\cite{kolesnikov2010image}. All the experimental runs were done on NVIDIA A100 GPU with $40$GB RAM, and the results are averaged over multiple iterations.
	
	\subsection{Dataset}
	In this work, we have used two dataset namely CIFAR-100 and CIFAR-20. CIFAR-100 is an image classification dataset comprising $60,000$ colour images, each $32\times32$ pixels, spanning $100$ classes. Each class includes $600$ images, with $500$ for training and $100$ for testing. The dataset is organized into $20$ super classes, sometimes called as CIFAR20 (formally known as CIFAR-100 coarse labels). For the machine unlearning experimets, we have constructed a forget set by randomly selecting $7$ and $5$  classes from CIFAR100 and CIFAR20 dataset respectively. For CIFAR100, $7$ forget classes are \textit{‘Rocket’, ‘Mushroom’, ‘Lamp’, ‘Apple’, ‘Baby’, ‘Sea’,} and \textit{‘Mountain’}. For CIFAR20 the forget classes are \textit{‘Electrical devices’, ‘Vehicles-2’, ‘Vegetables’, ‘People’} and \textit{‘Natural Scenes’}. 
	

	\begin{table}[!t]
	\small
	\centering
	\caption{Accuracy comparison of DFMU against SSD\cite{foster2024fast} for $f_{acc}$ and $r_{acc}$ for CIFAR-100 dataset. }
	\begin{tabular}{|c|c|c|c|c|c|c|c|c|}
		\hline
		\multirow{3}{*}{\textbf{Class}} & \multicolumn{2}{|c|}{\textbf{Base}}  &  \multicolumn{2}{|c|}{\textbf{SSD}} &\multicolumn{2}{|c|}{\textbf{DFMU}}  \\
		& \multicolumn{2}{|c|}{\textbf{Accuracy(\%)}}  &\multicolumn{2}{|c|}{\textbf{Accuracy(\%)}}&\multicolumn{2}{|c|}{\textbf{Accuracy(\%)}} \\ \cline{2-7}
		& $f_{acc}$& $r_{acc} $& $f_{acc}$ & $r_{acc}$ & $f_{acc}$ & $r_{acc}$ \\ \hline 
		Rocket & 93.70	& 92.66	& 0.00	& 0.72	& 0.00	& 79.35\\ \hline
		Mushroom & 99.22 & 92.59 & 0.00 & 41.11 & 0.00	& 86.75 \\ \hline
		Lamp & 97.66 & 92.61 & 0.00 & 1.29 & 0.78 & 82.95 \\ \hline
		Apple & 98.44 & 92.60 & 0.00 & 87.93 & 0.00 & 83.71 \\ \hline 
		Baby & 87.50 & 92.75 & 0.00  & 0.99 & 0.00 & 85.89 \\ \hline
		Sea & 87.51 & 92.67 & 0.00 & 0.74 & 2.34 & 79.90 \\ \hline 
		Mountain & 95.31 & 92.64  & 0.00 & 1.42  & 0.00	 & 80.75 \\ \hline 
	\end{tabular}
	\label{results_CIFAR100}
\end{table}

	\begin{table}[!t]
	\small
	\centering
	\caption{Accuracy comparison of DFMU against SSD\cite{foster2024fast} for $f_{acc}$ and $r_{acc}$ for CIFAR-20 dataset. }
	\begin{tabular}{|c|c|c|c|c|c|c|c|c|}
		\hline
		\multirow{3}{*}{\textbf{Class}} & \multicolumn{2}{|c|}{\textbf{Base}}  &  \multicolumn{2}{|c|}{\textbf{SSD}} &\multicolumn{2}{|c|}{\textbf{DFMU}}  \\
		& \multicolumn{2}{|c|}{\textbf{Accuracy(\%)}}  &\multicolumn{2}{|c|}{\textbf{Accuracy(\%)}}&\multicolumn{2}{|c|}{\textbf{Accuracy(\%)}} \\ \cline{2-7}
		& $f_{acc}$& $r_{acc} $& $f_{acc}$ & $r_{acc}$ & $f_{acc}$ & $r_{acc}$ \\ \hline 
		Elec\_Dev &96.78	& 95.84	& 0.00 &	6.99	& 0.39	& 84.25	\\ \hline		
		Vehicle2 &	95.42&	95.92	& 0.00	& 5.29	& 0.20	& 81.29	\\ \hline			
		Veg	& 97.76	&95.79&	0.00	& 18.57	& 7.81	& 91.56 \\ \hline
		People & 	97.42 &	95.81 &	0.00	& 4.88	& 0.39	& 87.86 \\ \hline
		Nat\_Scen & 	88.64&	96.27	& 0.00	& 5.26	& 4.84	& 85.12 \\ \hline
	\end{tabular}
	\label{results_CIFAR20}
\end{table}
	
	\begin{figure}[!t]
		\centering
		\includegraphics[width=8.5cm]{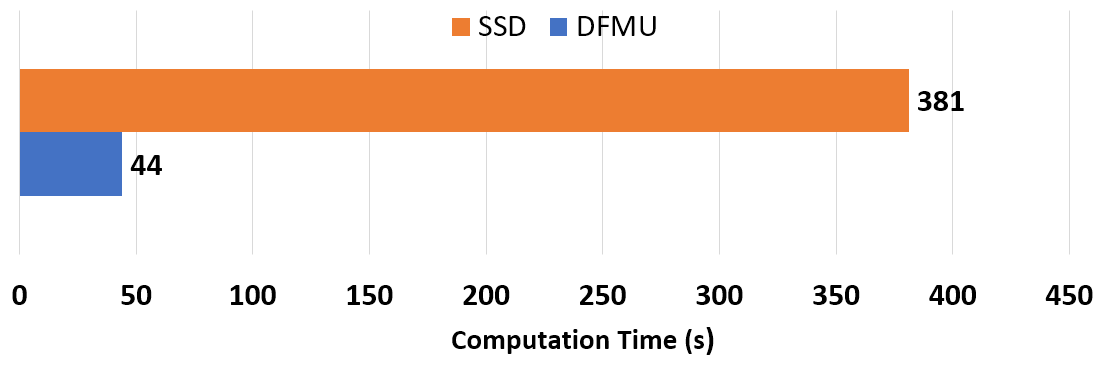}
		\caption{Average time taken for running the respective algorithm per class}
		\label{fig::computation_time}
	\end{figure}
	

	\subsection{Evaluation Metrics}
	Accuracy on the forget set ($f_{acc}$) and retain set ($r_{acc}$) is used to validate forgetting while retaining overall model performance. In a classification setting as here, if a model makes $N$ predictions and correctly classifies $C$ of them, then accuracy is given by:
	\begin{eqnarray}
		Accuracy = \frac{C}{N}\times 100 \%
	\end{eqnarray}
	
	\subsection{Results}
	
	\subsubsection{Classification Performance}
	For a fair comparison we are comparing the proposed method with SSD \cite{foster2024fast}. The results of the proposed method along with their comparison with the SSD \cite{foster2024fast} are summarized in Table \ref{results_CIFAR100} and Table \ref{results_CIFAR20}. Base accuracy is the accuracy of ViT model when trained on CIFAR-100 and CIFAR-20 datasets. The DFMU accuracy and SSD accuracy are the accuracies after applying the concept of machine unlearning. It is observed from Table \ref{results_CIFAR100} that, for CIFAR-100, DFMU has better retain accuracy compared to SSD and at the same time have competitive forget accuracy. Similarly, from Table \ref{results_CIFAR20}, it is observed that for CIFAR-20 dataset, DFMU obtain better retain accuracy and have competitive forget accuracy.

	Additionally, the time comparison between SSD and proposed method is shown in Fig. \ref{fig::computation_time}. It is evident that the average time taken for DFMU to process a single class is just $44s$ as compared to $381s$ in case of SSD.
	
	

	
	\subsubsection{Impact of data cardinality}
\begin{figure}[!t]
	\centering
	\includegraphics[width=8.5cm]{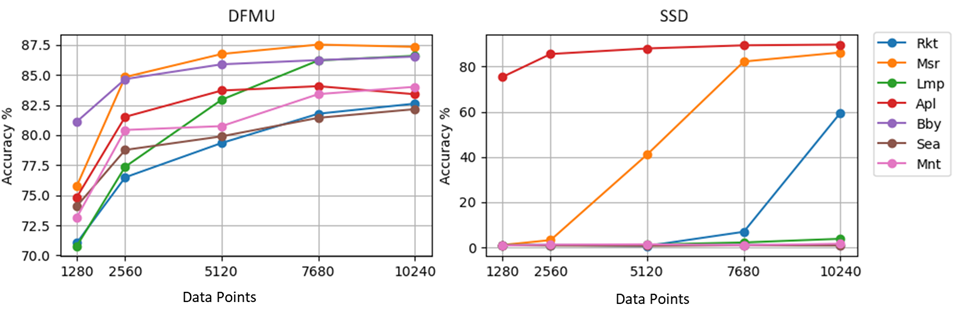}
	\caption{CIFAR-100 retain-dataset accuracy comparison of DFMU against SSD measured on different data cardinality}
	\label{fig::impact_data_cardinality2}
\end{figure}

\begin{figure}[!t]
	\centering
	\includegraphics[width=8.5cm]{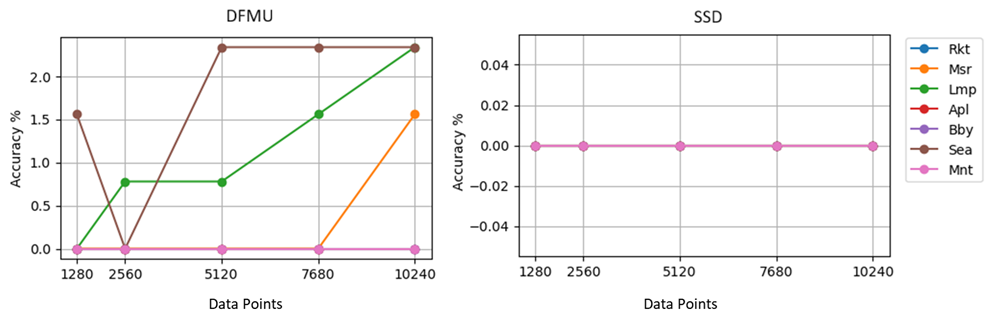}
	\caption{CIFAR-100 forget-dataset accuracy comparison of DFMU against SSD measured on different data cardinality}
	\label{fig::impact_data_cardinality3}
\end{figure}

\begin{figure}[!t]
	\centering
	\includegraphics[width=8.5cm]{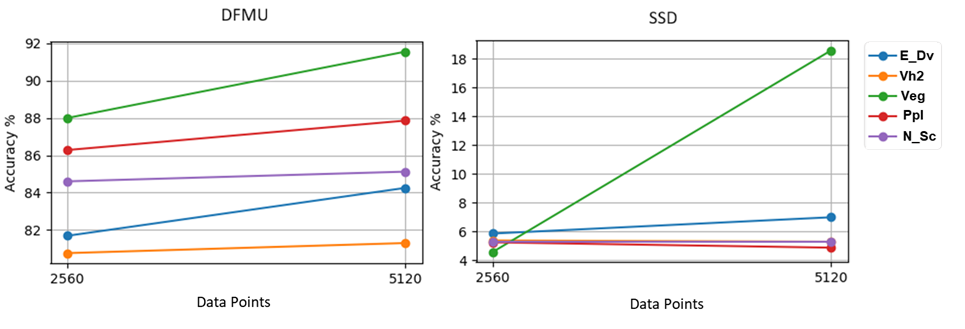}
	\caption{CIFAR-20 retain-dataset accuracy comparison of DFMU against SSD measured on different data cardinality}
	\label{fig::impact_data_cardinality2_2}
\end{figure}

\begin{figure}[!t]
	\centering
	\includegraphics[width=8.5cm]{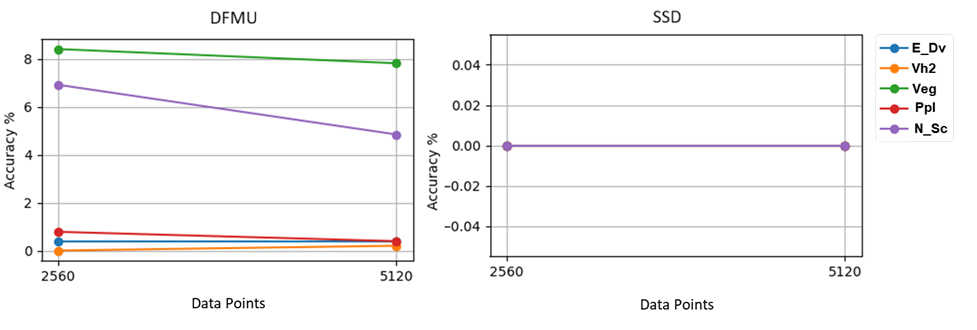}
	\caption{CIFAR-20 forget-dataset accuracy comparison of DFMU against SSD measured on different data cardinality}
	\label{fig::impact_data_cardinality3_2}
\end{figure}

	The amount of data samples available for calculating the importance scores is observed to have significant effect in retain-dataset accuracy of DFMU. When compared with SSD with the same number of data samples, DFMU outperforms SSD, starting from the beginning itself. Results of a set of experiments conducted on DFMU using different data samples is portrayed in Fig. \ref{fig::impact_data_cardinality2}, Fig. \ref{fig::impact_data_cardinality3}, Fig. \ref{fig::impact_data_cardinality2_2}, and Fig. \ref{fig::impact_data_cardinality3_2}. For CIFAR-100 dataset, we observed from  Fig. \ref{fig::impact_data_cardinality2} that the retain accuracy of DFMU increases for all categories but SSD convergence is comparatively slow. SSD only shows good convergence for category "Apple" which shows similar behaviour as DFMU. SSD struggles for convergence for most of the classes (4 out of 7), even after training of 10240 samples, "Mushroom", "Rocket" and "Apple" classes' retain accuracy is less than DFMU. Additionally, from Fig. \ref{fig::impact_data_cardinality3}, it can be observed that the forget accuracy of SSD is always "0". But, at the same time the retain accuracy of SSD is compromised, whereas our proposed method (DFMU) is achieving good retain accuracy with nominal rise in forget accuracy than 0\% in three classes ("Sea", "Lmap", and "Mushroom"). Hence, our method  shows better tradeoff between retain and forget accuracy compared to SSD method. Empirically it is analyzed that DFMU achieves $40\%$ more retain-accuracy with just $13\%$ of data samples as compared to SSD. As the required number of data samples are far less, DFMU reduces computational complexities and storage requirements considerably. 
	
	Similarly, CIFAR-20 dataset shows better tradeoff between retain and forget accuracy. From Fig. \ref{fig::impact_data_cardinality2_2} and Fig. \ref{fig::impact_data_cardinality3_2} it seems clearly that the proposed method's retain accuracy convergence is much better than SSD and the forget accuracy is competative. 

	
	\subsection{Ablation Study}
\begin{figure}[!t]
		\centering
		\includegraphics[width=8.5cm]{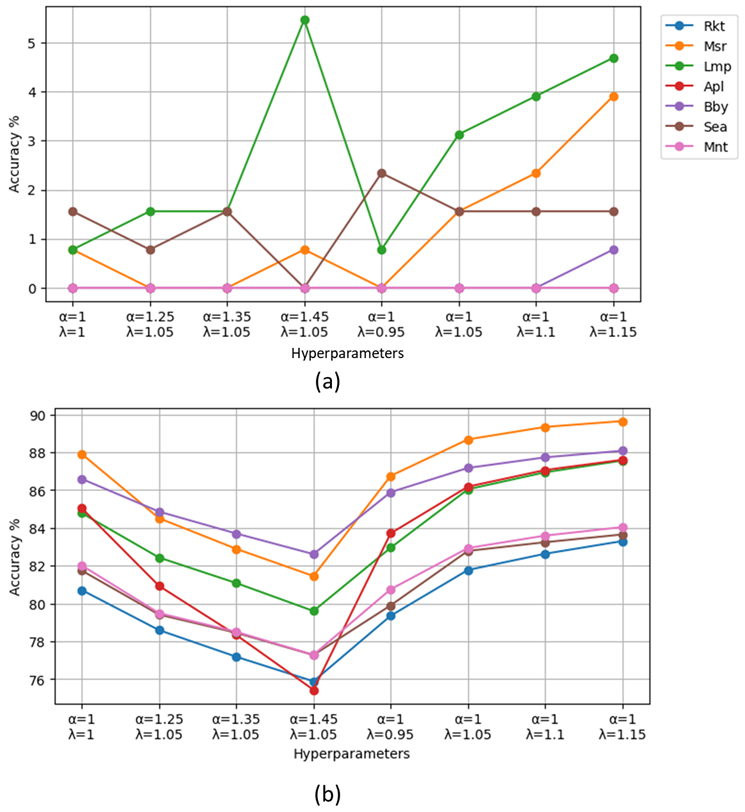}
		\caption{Impact of hyperparameter $\alpha$ and $\lambda$ on (a) Forget Accuracy and (b) Retain Accuracy for CIFAR-100 dataset}
		\label{fig::impact_data_cardinality5}
	\end{figure}

	For the ablation study, we conduct experiments to observe the impact of hyperparameters (scaling constant ($\alpha$) and perturb constant ($\lambda$)) on CIFAR-100 dataset. For the ablation, experiments are conducted on a series of values for $\alpha$ and $\lambda$.  Results of these experiments are given in Fig. \ref{fig::impact_data_cardinality5}. The best results are given by a combination of $\alpha$ = 1 and $\lambda$ = 0.95, since this values are giving good tradeoff between retain and forget accuracy of various classes. 

	\section{Conclusion}
	\label{sec:typestyle}
	
	In this work, Data-Frugal Machine Unlearning, a retraining-free approach is proposed for the task of machine unlearning that has a faster convergence rate for retain-dataset accuracy. The core of the proposed approach lies in the knowledge measurement of layers and the generation of importance scores signifying the saliency of neurons involved in the model’s knowledge about the forget-dataset and the complete-dataset. It is observed empirically that DFMU achieves more than $40\%$ more retain-accuracy with just $13\%$ of data samples in comparison with competitor method (SSD) on CIFAR-100 dataset. Further, DFMU also averages $88\%$ faster processing time against SSD for forgetting a given class.

	\bibliographystyle{IEEEbib}
	\bibliography{refs}
	
\end{document}